\documentclass[10pt,conference]{IEEEtran}
\usepackage{booktabs}
\usepackage{graphicx}
\usepackage{amsmath}
\usepackage{url}
\usepackage{xcolor}
\usepackage{microtype}
\usepackage[T1]{fontenc}
\usepackage{newtxtext,newtxmath}
\usepackage{tikz}
\IEEEoverridecommandlockouts

\title{One Word, Different Action: A Real-Robot Benchmark for Language-Conditioned Embodied Reasoning}

\author{%
\IEEEauthorblockN{Yiwei Liu\IEEEauthorrefmark{1}, \textit{Student Member, IEEE};\quad
Luwei Yang\IEEEauthorrefmark{2}, \textit{Member, IEEE};\quad
and Shunbo Lei\IEEEauthorrefmark{1}\IEEEauthorrefmark{2}, \textit{Senior Member, IEEE}}
\IEEEauthorblockA{\IEEEauthorrefmark{1}School of Science and Engineering, The Chinese University of Hong Kong, Shenzhen, Guangdong 518172, China}
\IEEEauthorblockA{\IEEEauthorrefmark{2}Shenzhen Research Institute of Big Data (SRIBD), Shenzhen 518172, China}
\IEEEauthorblockA{Email: yiweiliu1@link.cuhk.edu.cn; yangluwei@sribd.cn; leishunbo@cuhk.edu.cn}
\thanks{This work was supported in part by the National Natural Science Foundation of China under Grant 52307145, in part by the Shenzhen Basic Research Fund (Natural Science Foundation) under Grants QNXMB20250701091813018 and JCYJ20240813113532042, and in part by the Shenzhen Research Institute of Big Data (SRIBD) under Grant J00220250001. (Corresponding author: Shunbo Lei.)}
}

\IEEEaftertitletext{%
\vspace{-2.0\baselineskip}%
\refstepcounter{figure}%
\label{fig:decision-invariance-sensitivity}%
\noindent\hbox to \textwidth{\hss\includegraphics[width=\textwidth]{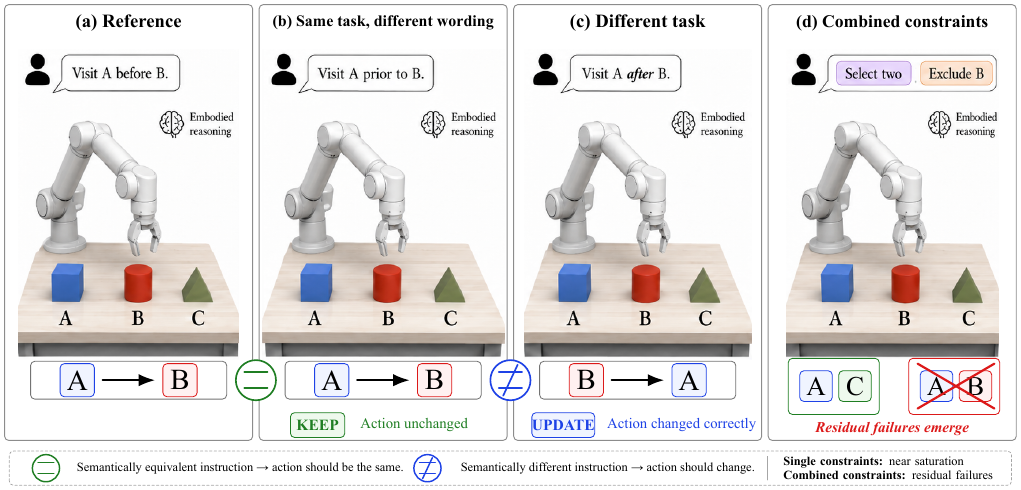}\hss}\endgraf
\vspace{-1pt}%
\noindent\parbox{\textwidth}{\centering\footnotesize
\textbf{Fig. \thefigure.}\ Illustration of decision invariance and decision sensitivity under a fixed physical decision state. The executable decision remains unchanged under the semantically equivalent reformulation in (a)--(b) and changes under the temporal flip in (c); combined constraints in (d) illustrate how residual failures can emerge.}%
\vspace{4pt}%
}

\begin{document}

\bstctlcite{BSTcontrol}
\maketitle

\begin{abstract}
Changes in natural-language instructions can directly alter the behavior ultimately executed by a robot, but such changes do not always imply that the task itself has changed. We propose One Word, Different Action, a language-conditioned executable decision benchmark based on real-robot physical decision anchors. It evaluates robot behavioral responses under task-preserving and task-changing conditions and examines joint reasoning over multiple task constraints. The benchmark uses decision invariance and decision sensitivity to measure action stability when task semantics are preserved and correct decision updates when task requirements change. Matched atomic constituents directly relate correct handling of individual constraints to the corresponding joint executable decision. Experiments show that atomic-task correctness does not sufficiently represent compositional reasoning ability: even when every constituent constraint is handled correctly, a model may still produce an erroneous final action decision on the corresponding compositional task. These results indicate that the key challenge in language-conditioned robot reasoning lies not only in understanding individual task requirements, but also in integrating multiple correctly handled constraints into consistent executable behavior.
\end{abstract}

\section{Introduction}

Natural language is becoming an important conditional input for robot decision making. Language models and multimodal foundation models can now connect high-level task descriptions with executable skills, visual states, and robot actions \cite{pmlr-v205-ichter23a,pmlr-v202-driess23a,pmlr-v229-zitkovich23a}. Language-conditioned manipulation has further expanded the range of tasks covered by this paradigm \cite{mees2022calvin,pmlr-v202-jiang23b}, while large-scale robot learning has continued to extend the connection from language-conditioned representations to executable robot actions \cite{openx24,pmlr-v270-kim25c}. As the link between language and physical execution becomes increasingly direct, the result of instruction understanding is ultimately manifested as a concrete robot behavior. A basic question therefore arises: when a user changes an instruction, should the robot preserve its current action or change it accordingly?

Existing embodied evaluations have expanded from overall task success rates to instruction understanding, perceptual reasoning, planning, and concrete manipulation decisions \cite{pmlr-v267-yang25f,luo2026robobenchcomprehensiveevaluationbenchmark,Zhang_2025_ICCV}, as well as low-level manipulation reasoning \cite{pmlr-v305-zhao25a}. At the same time, research on robot instructions has begun to systematically address user corrections, task switching, and language reformulations \cite{chiyah-garcia-etal-2024-repairs,stoyanchev-etal-2026-context,li2025switchvlaexecutionawaretaskswitching}; controlled paraphrase and minimal-pair evaluations further probe the stability and selectivity of language-conditioned decisions \cite{kim2026liberoparadiagnosticbenchmarkmetrics,oba-sugawara-2026-cxmp}. Compositional generalization research examines whether familiar task components, atomic skills, and linguistic factors can form correct behavior under new joint conditions \cite{mesa2026,qi2026scalestrategicallylearningcompositional,wu2026atombench}; diagnostic studies of sequential robot tasks and explicit-constraint methods further connect complex language requirements to verifiable robot decision conditions \cite{wang2026diagnosingcompositional,quartey2025verifiably}. Together, these studies establish evaluation foundations for language variation and task composition.

What remains missing is a direct correspondence between the two within the same real physical decision state. A change in an instruction may only alter its expression, or it may genuinely change the task requirements; multiple constraints may also act simultaneously on the same final action. If one considers only independent task accuracy or overall composite accuracy, it is impossible to determine whether an error on a joint task arises because a constituent constraint was not solved in the first place, or because constraints that had already been solved separately were not correctly composed into an executable decision.

We propose One Word, Different Action, a language-conditioned executable decision benchmark based on real-robot physical decision anchors. Each anchor fixes the physical state, candidate identities, and executable action space, and constructs task-preserving, task-changing, atomic, and compositional tasks within the same decision space. Decision invariance measures whether the action is correctly preserved under a task-preserving change, whereas decision sensitivity measures whether the action is correctly updated under a task-changing change; for compositional tasks, matched atomic constituents further establish a direct correspondence between constituent competence and joint compositional competence. Conditional composition success evaluates the corresponding joint task only when all matched atomic constituents have already been solved correctly, thereby directly testing whether constraints handled correctly in isolation can form a correct joint executable decision.

The experiments show that correct decisions on atomic tasks do not sufficiently characterize joint decision ability on compositional tasks. Even when all matched atomic constituents are correct, multiple models still produce an erroneous executable decision on the corresponding compositional task; at the same time, correctly updating the decision after a task change is overall more difficult than correctly preserving the decision under task preservation. These results indicate that a key challenge in language-conditioned robot reasoning lies not only in understanding individual task constraints, but also in integrating multiple constraints that have already been handled correctly into consistent robot behavior.

\section{Related Work}

\subsection*{A.\quad Embodied Reasoning and Language-Conditioned Robot Decision Making}

Existing embodied-intelligence benchmarks increasingly move beyond final task success toward fine-grained evaluation of language understanding, task reasoning, and robot decision making. RoboBench evaluates multimodal foundation models as embodied decision makers across instruction understanding, perceptual reasoning, planning, and action selection \cite{luo2026robobenchcomprehensiveevaluationbenchmark}. VLABench studies language-conditioned manipulation through complex natural-language tasks and long-horizon behavior, while ManipBench focuses more closely on low-level manipulation reasoning involving object interactions, action relations, and operational processes \cite{Zhang_2025_ICCV,pmlr-v305-zhao25a}. Together, these benchmarks establish increasingly diagnostic evaluations of language-conditioned robot behavior.

Our benchmark focuses on language changes within the same physical decision state. A physical decision anchor fixes the physical state, candidate identities, and executable action space, while different instructions alter the task constraints imposed on that space. Task-preserving changes, task-changing changes, and multi-constraint compositions can therefore be evaluated under a common executable decision representation. This organization directly relates changes in language constraints to changes in robot action decisions within the same physical substrate.

\subsection*{B.\quad Instruction Changes, Corrections, and Decision Updates}

A complementary line of work studies how robots respond to instruction revisions, corrections, and alternative linguistic formulations. BlockWorld-Repairs evaluates whether multimodal models can recover from an incorrect task interpretation using subsequent user corrections, while Stoyanchev et al. study corrections, refinements, and interruptions in human-robot dialogue and how language models use interaction context to update robot behavior \cite{chiyah-garcia-etal-2024-repairs,stoyanchev-etal-2026-context}. SwitchVLA addresses task switching during execution by adapting robot behavior to the current execution state and a new task instruction \cite{li2025switchvlaexecutionawaretaskswitching}. For task-preserving language variation, LIBERO-Para evaluates the stability of VLA policies under paraphrases of the same manipulation task, while CxMP uses controlled linguistic minimal pairs to test whether models respond selectively to meaning-bearing changes in language \cite{kim2026liberoparadiagnosticbenchmarkmetrics,oba-sugawara-2026-cxmp}.

Our benchmark treats preservation and update as complementary outcomes of the same executable decision problem. Decision invariance measures whether the correct action is preserved when task semantics remain unchanged, whereas decision sensitivity measures whether the executable decision is updated when the task constraints change. Both relations are defined over the same physical decision anchor and candidate action space, allowing the effect of an instruction change to be measured directly through whether the corresponding robot decision should remain fixed or change.

\subsection*{C.\quad Compositional Generalization and Multi-Constraint Robot Reasoning}

Compositional generalization studies whether familiar task components can be recombined into correct robot behavior under new joint conditions. MESA evaluates unseen subtask compositions alongside spatial and semantic generalization in language-conditioned manipulation \cite{mesa2026}. Qi et al.~\cite{qi2026scalestrategicallylearningcompositional} study instruction-factor bias associated with factors such as objects, actions, attributes, and spatial relations under compositional generalization. ATOM-Bench explicitly separates atomic skill acquisition from compositional reuse in real-world manipulation policies, and Wang et al.~\cite{wang2026diagnosingcompositional} study compositional behavior through instruction-space coverage and the recombination of familiar instruction components. LIMP approaches complex instruction following through explicit symbolic constraints, using jointly satisfied spatial and temporal requirements to determine executable robot behavior \cite{quartey2025verifiably}.

Our benchmark directly aligns atomic competence with compositional competence within a fixed real-robot decision space. Each compositional task is paired with matched atomic constituents that share the same physical decision anchor, candidate set, and constituent constraint semantics. Conditional composition success evaluates the joint task only after all matched atomic constituents have been solved correctly, directly testing whether individually correct constraint decisions compose into the correct joint executable decision.

\section{Language-Conditioned Executable Decision Reasoning}

\subsection{Physical Decision State and Executable Action Space}
Consider a robot in a physical decision state $s$. This state contains the physical information required for the current decision and determines a finite executable action space
\begin{equation}
    \mathcal{A}(s)=\{a_1,a_2,\ldots,a_N\}.
\end{equation}
Each $a_i$ denotes a candidate action with a well-defined physical meaning. Depending on the task, a candidate may be a target-object selection, a set of actions, or an action sequence with a prescribed execution order. We study the formation of a correct executable decision within a fixed physical state and candidate action space, rather than the generation of arbitrary robot behavior from an open action space. We refer to this fixed physical state together with its finite candidate action space as a \emph{physical decision anchor}.

Given a natural-language instruction $x$, the model produces a decision $\pi(x,s)$. For each task instance, the physical state, candidate action space, and task constraints jointly determine a unique correct decision $a^*(x,s)$. Language reasoning therefore terminates in an executable choice in the robot action space rather than an independent semantic label. A prediction is correct only when its final action, action set, or action sequence exactly matches $a^*(x,s)$.

This representation directly connects language understanding to robot behavior. In a controlled comparison, $s$ and $\mathcal{A}(s)$ remain fixed, while the instruction determines which elements of the action space enter the final decision and which relations they must satisfy. Instruction changes can consequently be characterized directly by whether the correct executable decision changes.

\subsection{Decision Invariance and Decision Sensitivity}
Language changes under a fixed physical state do not always require a change in robot behavior. Let $x^+$ denote a task-preserving transformation of $x$. When the two expressions specify the same task, their correct decisions satisfy
\begin{equation}
    a^*(x,s)=a^*(x^+,s).
\end{equation}
The robot should retain the same correct executable decision under the two expressions. We call this capability \emph{decision invariance}. It requires stability under language changes that leave the task requirements unchanged, rather than merely producing two textually similar outputs.

Let $x^-$ denote a task-changing transformation. When the instruction change modifies a task requirement that affects execution,
\begin{equation}
    a^*(x,s)\ne a^*(x^-,s).
\end{equation}
The robot must update its executable decision accordingly. We call this capability \emph{decision sensitivity}. It requires the model to identify a linguistic difference that changes the behavioral requirement and to reflect that difference in its action selection.

Figure~\ref{fig:decision-invariance-sensitivity} illustrates these two relations. With the physical state fixed, a task-preserving change in expression corresponds to the same decision, whereas a change in the task requirement updates the correct decision. Together, the two properties characterize selective stability under language conditioning: the robot must preserve correct behavior when the task is preserved and revise it correctly when the task changes.

\subsection{Atomic Task Constraints}
Language tasks act on the executable action space through four atomic operator types: \emph{action inhibition} ($N$), \emph{set exclusion} ($E$), \emph{cardinality} ($Q$), and \emph{temporal ordering} ($T$).

$N$ removes or inhibits a candidate action or object from the final decision. $E$ excludes specified members from the candidate set. $Q$ specifies the required number of selected candidates. $T$ specifies the required ordering among candidate actions and thereby determines an ordered action sequence.

\begin{figure*}[t]
\centering
\input{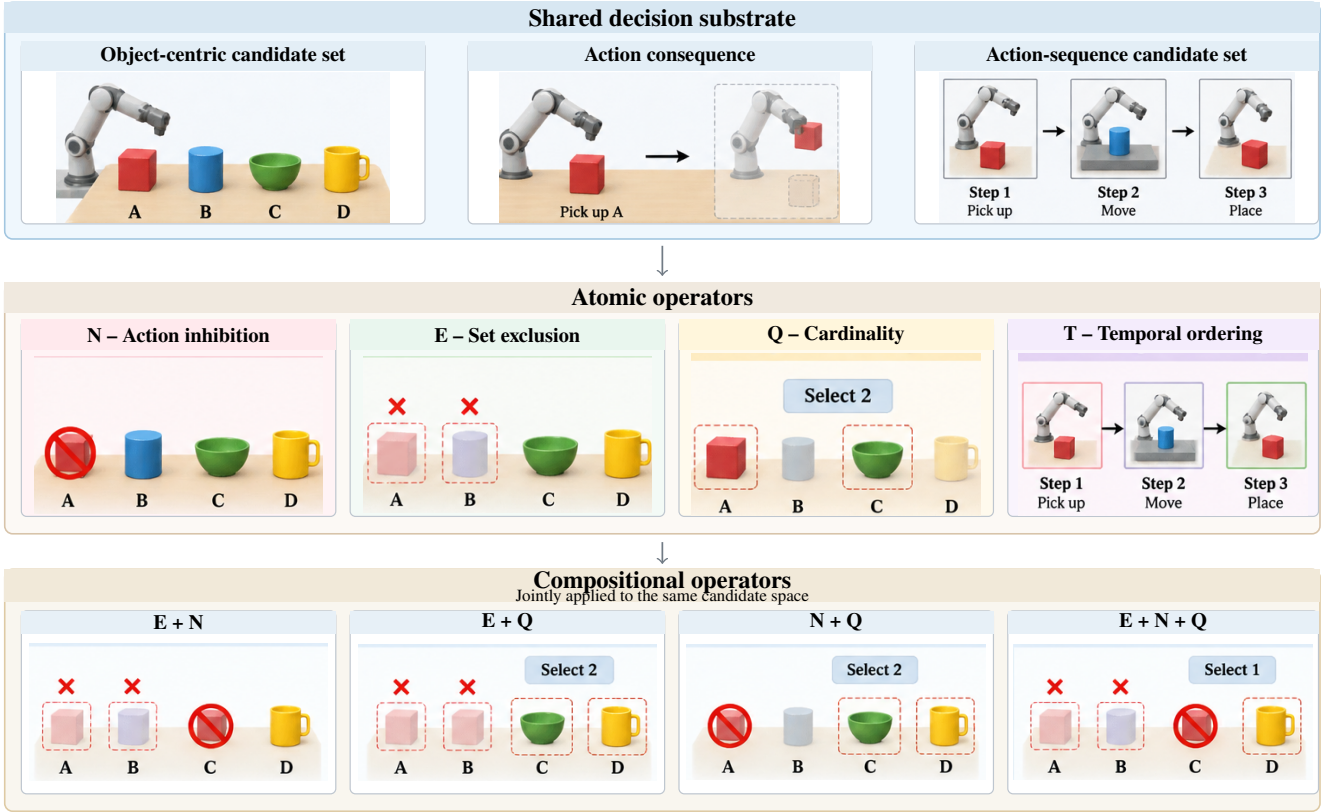}
\caption{Atomic operators and their composition over a shared executable action space. $N$, $E$, $Q$, and $T$ denote action inhibition, set exclusion, cardinality, and temporal ordering, respectively. The current object-centric compositional subset jointly applies $N$, $E$, and $Q$ as $E+N$, $E+Q$, $N+Q$, and $E+N+Q$, while $T$ is evaluated on atomic action-sequence tasks.}
\label{fig:atomic-operator-composition}
\end{figure*}

These constraints act on different decision structures but share the same basic form: the physical state and candidate action space are fixed, and the task constraint determines which actions constitute the final executable decision. Atomic tasks thus provide a direct evaluation of the mapping from an individual language condition to a robot action decision. Figure~\ref{fig:atomic-operator-composition} summarizes these operators and their supported compositional labels.

\subsection{Compositional Task Constraints}
A compositional task applies two or more atomic operators simultaneously to the same candidate action space. The final executable decision must satisfy all active operators simultaneously.

The current compositional subset applies $N$, $E$, and $Q$ on object-centric anchors. It comprises set exclusion plus action inhibition ($E+N$), set exclusion plus cardinality ($E+Q$), action inhibition plus cardinality ($N+Q$), and set exclusion, action inhibition, and cardinality ($E+N+Q$). $T$ is evaluated as an atomic temporal-ordering operator on action-sequence anchors and is not part of the current compositional subset.

We distinguish \emph{atomic competence}, correctness on an atomic task, from \emph{compositional competence}, correctness on the corresponding joint task. A model may solve the corresponding atomic tasks separately and still fail to produce a correct joint executable decision. Compositional reasoning evaluates whether individually solved local conditions can be integrated into one consistent robot decision.

\subsection{Matched-Constituent Evaluation}
Each compositional task has atomic tasks corresponding to its constituent operators, termed \emph{matched atomic constituents}. An $E+Q$ task has the corresponding $E$ and $Q$ atomic tasks, whereas an $E+N+Q$ task has the corresponding $E$, $N$, and $Q$ atomic tasks. Compositional and atomic tasks share the same physical decision state, candidate identities, and executable action space; they differ in whether the operators are applied independently or jointly.

Conditional composition evaluation asks whether a compositional task is solved correctly after all of its matched atomic constituent tasks have already been solved correctly. It tests whether operators that are handled correctly in isolation can be integrated into a correct joint executable decision.

\section{Real-Robot Benchmark Construction}

\subsection{Physical Decision Anchors}
The benchmark is constructed from physical states collected and verified with a real robot. It contains 85 physical decision anchors: 21 object-centric anchors and 64 action-sequence anchors. Object-centric anchors provide explicit candidate objects for action-inhibition, set-exclusion, and cardinality tasks. Action-sequence anchors provide verified candidate actions and their relations for temporal-ordering tasks.

Within each anchor, candidate identities and executable semantics remain fixed. The candidate space supports target-object selections, object sets, or ordered action sequences, according to the anchor type.

\subsection{Task-Family Construction}
The atomic component contains 988 complete task families and 3,952 instances across the four atomic operator labels $N$, $E$, $Q$, and $T$. Each family applies one primary task constraint and includes task-preserving and task-changing transformations.

The compositional component uses 21 object-centric anchors and contains 102 complete families with 408 instances. It includes 82 two-constraint ($K=2$) families with 328 instances and 20 three-constraint ($K=3$) families with 80 instances. The supported composition types are $E+N$, $E+Q$, $N+Q$, and $E+N+Q$. Table~\ref{tab:representative-compositional-instructions} gives representative instructions sampled directly from the evaluated compositional benchmark.

\begin{table}[t]
\centering
\caption{Representative compositional instructions sampled from the evaluated benchmark.}
\label{tab:representative-compositional-instructions}
\footnotesize
\setlength{\tabcolsep}{3.5pt}
\renewcommand{\arraystretch}{1.10}
\begin{tabular}{@{}l p{0.84\columnwidth}@{}}
\toprule
Type & Representative instruction \\
\midrule
$E{+}N$ & ``Inspect all listed objects excluding object\_2; do not inspect object\_3.'' \\
$E{+}Q$ & ``Inspect exactly the following 2 objects: object\_3 and object\_4, excluding object\_2.'' \\
$N{+}Q$ & ``Inspect exactly the following two objects: object\_2 and object\_3, but do not inspect object\_4.'' \\
$E{+}N{+}Q$ & ``Choose exactly 1 object: object\_4; exclude object\_2 and never inspect object\_3.'' \\
\bottomrule
\end{tabular}

\end{table}

For each compositional family, corresponding atomic constituent tasks are constructed from the same physical decision anchor and candidate space, with each constituent applied separately. These matched atomic constituents support subsequent conditional evaluation.

\subsection{Ground Truth and Validation}
Underlying symbolic constraints are specified before their natural-language realizations. For each instance, the anchor, candidate space, and active constraints determine one unique legal executable decision.

Validation checks the physical decision anchor and candidate catalog, constraint consistency, and uniqueness of the executable decision. It also verifies membership, exclusion, cardinality, ordering, and joint constraint satisfaction.

\section{Evaluation Protocol}

\subsection{Models and Inference Configuration}
We evaluate four language models: GLM-5-FP8~\cite{glm5team2026glm5vibecodingagentic}, Qwen2.5-7B-Instruct~\cite{qwen2025qwen25technicalreport}, Mistral-7B-Instruct-v0.3~\cite{mistralai2024mistral7binstructv03}, and OLMo-2-1124-7B-Instruct~\cite{teamolmo2025olmo2furious}. All models receive the same task information, candidate decision space, and target output semantics, with formatting adapted only to the corresponding native instruction interface. Each model returns a machine-readable executable decision without an explanation.

All models use temperature 0, a maximum of 512 new tokens, and disabled sampling. No chain-of-thought explanation is requested or evaluated. The model identifiers and interfaces used in the evaluation are listed in Table~\ref{tab:model-config}.

\begin{table*}[t]
\centering
\caption{Model and decoding configuration.}
\label{tab:model-config}
\footnotesize
\setlength{\tabcolsep}{5pt}
\renewcommand{\arraystretch}{1.12}
\begin{tabular}{@{}lllcccl@{}}
\toprule
Model & Checkpoint identifier & Modality & Interface & Temperature & Max new tokens & Sampling \\
\midrule
GLM-5-FP8 & GLM-5-FP8 & Text & Local inference API & 0 & 512 & Disabled \\
Qwen2.5-7B-Instruct & Qwen/Qwen2.5-7B-Instruct & Text & Native instruct & 0 & 512 & Disabled \\
Mistral-7B-Instruct-v0.3 & mistralai/Mistral-7B-Instruct-v0.3 & Text & Native instruct & 0 & 512 & Disabled \\
OLMo-2-1124-7B-Instruct & allenai/OLMo-2-1124-7B-Instruct & Text & Native instruct & 0 & 512 & Disabled \\
\bottomrule
\end{tabular}

\end{table*}

\subsection{Exact Executable Decision Evaluation}
We report \emph{exact executable-decision accuracy}, which compares each model output directly with deterministic ground truth. Outputs that cannot be mapped to a valid executable decision are scored as incorrect. For a single-action task, the action must match exactly; for a set-valued task, membership must match exactly; and for a sequence-valued task, both membership and order must match exactly.

\subsection{Decision Invariance and Decision Sensitivity}
For a task-preserving pair, decision invariance requires both decisions to be correct and their normalized executable decisions to be identical. For a task-changing pair, decision sensitivity requires both decisions to be correct and the normalized executable decisions to differ as required by the changed constraint.

\subsection{Conditional Composition Success}
\emph{Conditional composition success} is the proportion of compositional tasks solved correctly among those for which all matched atomic constituent tasks are solved correctly. CCS measures compositional correctness after constituent competence has been established.

\subsection{Composition Structure}
Composition is analyzed by constraint depth, with $K=2$ and $K=3$, and by composition type, with $E+N$, $E+Q$, $N+Q$, and $E+N+Q$. These dimensions describe the structure of the compositional task and are evaluated using the same exact executable-decision protocol.

\section{Results}

\subsection{Benchmark Results}
Table~\ref{tab:main-results} reports atomic accuracy, compositional accuracy, conditional composition success, decision invariance, and decision sensitivity across the four evaluated models. Atomic accuracy is high for several models, whereas compositional accuracy is lower for all models.

\begin{table*}[t]
\centering
\caption{Atomic and compositional accuracy, conditional composition success, decision invariance, and decision sensitivity.}
\label{tab:main-results}
\footnotesize
\setlength{\tabcolsep}{6pt}
\begin{tabular}{lrrrrr}
\toprule
Model & Atomic & Composite & Cond. Comp. & Invariance & Sensitivity \\
\midrule
GLM-5-FP8 & 0.994 & 0.949 & 0.818 & 0.961 & 0.843 \\
Qwen2.5-7B-Instruct & 0.844 & 0.645 & 1.000 & 0.657 & 0.471 \\
Mistral-7B-Instruct-v0.3 & 0.978 & 0.836 & 0.610 & 0.755 & 0.696 \\
OLMo-2-1124-7B-Instruct & 0.878 & 0.828 & 0.929 & 0.814 & 0.667 \\
\bottomrule
\end{tabular}

\end{table*}

Conditional composition success evaluates the joint task after all matched atomic constituents are correct. GLM-5-FP8, Mistral-7B-Instruct-v0.3, and OLMo-2-1124-7B-Instruct retain conditional composition failures, whereas Qwen2.5-7B-Instruct is successful on all eligible compositional families.

Across all four models, decision invariance is higher than decision sensitivity. This pattern indicates more reliable preservation under task-preserving changes than revision under task-changing changes.

Table~\ref{tab:depth-failures} shows that GLM-5-FP8 is higher at $K=3$ than at $K=2$, whereas the other three models are lower at $K=3$. Table~\ref{tab:composition-types} shows that $E+Q$ is the strongest type for Qwen2.5-7B-Instruct, Mistral-7B-Instruct-v0.3, and OLMo-2-1124-7B-Instruct, whereas GLM-5-FP8 performs best on $E+Q$ and $E+N+Q$. The $E+N+Q$ type is comparatively weak for Mistral-7B-Instruct-v0.3 and OLMo-2-1124-7B-Instruct, while GLM-5-FP8 remains perfect on that type.

\begin{table}[t]
\centering
\caption{Composition-depth performance and conditional composition success.}
\label{tab:depth-failures}
\footnotesize
\setlength{\tabcolsep}{6pt}
\begin{tabular}{@{}lrrr@{}}
\toprule
Model & K=2 & K=3 & Cond. Comp. \\
\midrule
GLM-5-FP8 & 0.936 & 1.000 & 0.818 \\
Qwen2.5-7B & 0.720 & 0.338 & 1.000 \\
Mistral-7B & 0.890 & 0.613 & 0.610 \\
OLMo-2-7B & 0.948 & 0.338 & 0.929 \\
\bottomrule
\end{tabular}

\end{table}

\begin{table}[t]
\centering
\caption{Accuracy by compositional constraint type.}
\label{tab:composition-types}
\footnotesize
\setlength{\tabcolsep}{6pt}
\begin{tabular}{@{}lrrrr@{}}
\toprule
Model & E+N & E+Q & N+Q & E+N+Q \\
\midrule
GLM-5-FP8 & 0.925 & 1.000 & 0.911 & 1.000 \\
Qwen2.5-7B & 0.050 & 1.000 & 0.905 & 0.338 \\
Mistral-7B & 0.763 & 0.963 & 0.917 & 0.613 \\
OLMo-2-7B & 0.963 & 0.975 & 0.929 & 0.338 \\
\bottomrule
\end{tabular}

\end{table}

\subsection{Significance for Robot Decision Making}
Conditional composition success shows that atomic-task correctness is insufficient to establish joint compositional competence. Conditional composition errors remain after every constituent constraint has been handled correctly in isolation, so an evaluation limited to isolated constraints can overstate compositional reliability.

The consistent gap between decision invariance and decision sensitivity distinguishes preserving a correct executable decision from revising it after a genuine task change. Action preservation is more reliable under task-preserving changes, whereas task updates more readily expose errors in the executable decision.

Because benchmark outputs are executable decisions, composition failures correspond to an incorrect selected object or object set, an incorrect cardinality, failure to exclude a prohibited candidate, an incorrect action update, or an incorrect ordered action sequence where applicable. A model can handle each task requirement separately yet still fail to produce the joint executable decision required for robot behavior. Constituent competence therefore cannot substitute for joint executable verification.

\section{Future Work}

The current results show that correctly solving multiple atomic task constraints separately does not guarantee a correct joint executable decision. A direct direction for future work is to incorporate this gap explicitly into fine-tuning objectives. Training data could be organized around matched atomic constituents and their corresponding compositional tasks, so that the model receives supervision on both the correct decision for each individual constraint and the target behavior under their conjunction. This training signal would directly connect constituent-level competence to the joint executable decision instead of improving overall accuracy merely by increasing the number of independent instruction examples.

Further training could use a curriculum that progressively increases compositional complexity from atomic tasks to multi-constraint tasks, together with focused sampling of tasks for which all constituent constraints are correctly solved but the joint decision remains incorrect. Task-preserving and task-changing pairs could also be used jointly for fine-tuning, allowing the model to improve decision sensitivity to genuine task changes while maintaining decision invariance. Subsequent work can investigate whether different compositional supervision, training curricula, and hard-example sampling strategies improve conditional composition success, and whether such improvements reduce joint executable-decision errors while preserving atomic competence.

\section{Conclusion}

This work studies how language changes affect robot behavior from the perspective of real-robot executable decisions. By fixing a physical decision anchor, candidate identities, and executable action space, it places task-preserving changes, task-changing changes, and multi-constraint compositions within a single evaluation framework. Decision invariance and decision sensitivity describe the two basic capabilities of preserving or updating actions under language changes, while matched atomic constituents further separate compositional reasoning from overall task difficulty, enabling constituent competence and joint compositional competence to be directly compared within the same physical decision space. The experiments reveal that correctly solving individual task constraints does not mean that these constraints can naturally form a correct final decision under joint conditions; meanwhile, action updates after a genuine change in task requirements more readily expose errors than decision preservation under task-preserving conditions. Reliable language-conditioned robot decision making therefore requires not only correct responses to local constraints, but also preservation of their semantic relations when multiple constraints act simultaneously and consistent realization of those relations in a unique joint executable decision.

\bibliographystyle{IEEEtran}
\bibliography{references}

@IEEEtranBSTCTL{BSTcontrol,
  CTLuse_forced_etal       = "yes",
  CTLmax_names_forced_etal = "6",
  CTLnames_show_etal       = "1"
}

@misc{luo2026robobenchcomprehensiveevaluationbenchmark,
  title = {{RoboBench}: A comprehensive evaluation benchmark for multimodal large language models as embodied brain},
  author = {Luo, Yulin and Fan, Chun-Kai and Dong, Menghang and Shi, Jiayu and Mi, Xiangju and Zhao, Mengdi and Zhang, Bo-Wen and Chi, Cheng and Liu, Jiaming and Dai, Gaole and Zhang, Rongyu and An, Ruichuan and Wu, Kun and Che, Zhengping and Xie, Shaoxuan and Yao, Guocai and Zhao, Zhongxia and Wang, Pengwei and Liu, Guang and Wang, Zhongyuan and Huang, Tiejun and Zhang, Shanghang},
  year = {2026},
  eprint = {2510.17801},
  archivePrefix = {arXiv},
  primaryClass = {cs.RO},
  note = {arXiv:2510.17801}
}

@InProceedings{pmlr-v267-yang25f,
  title = {{EmbodiedBench}: Comprehensive benchmarking multi-modal large language models for vision-driven embodied agents},
  author = {Yang, Rui and Chen, Hanyang and Zhang, Junyu and Zhao, Mark and Qian, Cheng and Wang, Kangrui and Wang, Qineng and Koripella, Teja Venkat and Movahedi, Marziyeh and Li, Manling and Ji, Heng and Zhang, Huan and Zhang, Tong},
  booktitle = {Proceedings of the 42nd International Conference on Machine Learning},
  pages = {70576--70631},
  year = {2025},
  volume = {267},
  series = {Proceedings of Machine Learning Research},
  publisher = {PMLR},
}

@article{mees2022calvin,
  author = {Mees, Oier and Hermann, Lukas and Rosete-Beas, Erick and Burgard, Wolfram},
  title = {{CALVIN}: A benchmark for language-conditioned policy learning for long-horizon robot manipulation tasks},
  journal = {IEEE Robotics and Automation Letters},
  volume = {7},
  number = {3},
  pages = {7327--7334},
  year = {2022},
  doi = {10.1109/LRA.2022.3180108}
}

@InProceedings{pmlr-v202-jiang23b,
  title = {{VIMA}: Robot manipulation with multimodal prompts},
  author = {Jiang, Yunfan and Gupta, Agrim and Zhang, Zichen and Wang, Guanzhi and Dou, Yongqiang and Chen, Yanjun and Fei-Fei, Li and Anandkumar, Anima and Zhu, Yuke and Fan, Linxi},
  booktitle = {Proceedings of the 40th International Conference on Machine Learning},
  pages = {14975--15022},
  year = {2023},
  volume = {202},
  series = {Proceedings of Machine Learning Research},
  publisher = {PMLR}
}

@InProceedings{pmlr-v205-ichter23a,
  title = {Do as {I} can, not as {I} say: Grounding language in robotic affordances},
  author = {Ichter, Brian and Brohan, Anthony and Chebotar, Yevgen and Finn, Chelsea and Hausman, Karol and Herzog, Alexander and Ho, Daniel and Ibarz, Julian and Irpan, Alex and Jang, Eric and Julian, Ryan and Kalashnikov, Dmitry and Levine, Sergey and Lu, Yao and Parada, Carolina and Rao, Kanishka and Sermanet, Pierre and Toshev, Alexander T and Vanhoucke, Vincent and Xia, Fei and Xiao, Ted and Xu, Peng and Yan, Mengyuan and Brown, Noah and Ahn, Michael and Cortes, Omar and Sievers, Nicolas and Tan, Clayton and Xu, Sichun and Reyes, Diego and Rettinghouse, Jarek and Quiambao, Jornell and Pastor, Peter and Luu, Linda and Lee, Kuang-Huei and Kuang, Yuheng and Jesmonth, Sally and Joshi, Nikhil J. and Jeffrey, Kyle and Ruano, Rosario Jauregui and Hsu, Jasmine and Gopalakrishnan, Keerthana and David, Byron and Zeng, Andy and Fu, Chuyuan Kelly},
  booktitle = {Proceedings of The 6th Conference on Robot Learning},
  pages = {287--318},
  year = {2023},
  volume = {205},
  series = {Proceedings of Machine Learning Research},
  publisher = {PMLR}
}

@InProceedings{pmlr-v202-driess23a,
  title = {{PaLM-E}: An embodied multimodal language model},
  author = {Driess, Danny and Xia, Fei and Sajjadi, Mehdi S. M. and Lynch, Corey and Chowdhery, Aakanksha and Ichter, Brian and Wahid, Ayzaan and Tompson, Jonathan and Vuong, Quan and Yu, Tianhe and Huang, Wenlong and Chebotar, Yevgen and Sermanet, Pierre and Duckworth, Daniel and Levine, Sergey and Vanhoucke, Vincent and Hausman, Karol and Toussaint, Marc and Greff, Klaus and Zeng, Andy and Mordatch, Igor and Florence, Pete},
  booktitle = {Proceedings of the 40th International Conference on Machine Learning},
  pages = {8469--8488},
  year = {2023},
  volume = {202},
  series = {Proceedings of Machine Learning Research},
  publisher = {PMLR}
}

@InProceedings{pmlr-v229-zitkovich23a,
  title = {{RT-2}: Vision-language-action models transfer web knowledge to robotic control},
  author = {Zitkovich, Brianna and Yu, Tianhe and Xu, Sichun and Xu, Peng and Xiao, Ted and Xia, Fei and Wu, Jialin and Wohlhart, Paul and Welker, Stefan and Wahid, Ayzaan and Vuong, Quan and Vanhoucke, Vincent and Tran, Huong and Soricut, Radu and Singh, Anikait and Singh, Jaspiar and Sermanet, Pierre and Sanketi, Pannag R. and Salazar, Grecia and Ryoo, Michael S. and Reymann, Krista and Rao, Kanishka and Pertsch, Karl and Mordatch, Igor and Michalewski, Henryk and Lu, Yao and Levine, Sergey and Lee, Lisa and Lee, Tsang-Wei Edward and Leal, Isabel and Kuang, Yuheng and Kalashnikov, Dmitry and Julian, Ryan and Joshi, Nikhil J. and Irpan, Alex and Ichter, Brian and Hsu, Jasmine and Herzog, Alexander and Hausman, Karol and Gopalakrishnan, Keerthana and Fu, Chuyuan and Florence, Pete and Finn, Chelsea and Dubey, Kumar Avinava and Driess, Danny and Ding, Tianli and Choromanski, Krzysztof Marcin and Chen, Xi and Chebotar, Yevgen and Carbajal, Justice and Brown, Noah and Brohan, Anthony and Arenas, Montserrat Gonzalez and Han, Kehang},
  booktitle = {Proceedings of The 7th Conference on Robot Learning},
  pages = {2165--2183},
  year = {2023},
  volume = {229},
  series = {Proceedings of Machine Learning Research},
  publisher = {PMLR}
}

@inproceedings{openx24,
  title = {{Open X-Embodiment}: Robotic learning datasets and {RT-X} models},
  author = {{Open X-Embodiment Collaboration}},
  booktitle = {2024 IEEE International Conference on Robotics and Automation (ICRA)},
  pages = {6892--6903},
  year = {2024},
  publisher = {IEEE},
  doi = {10.1109/ICRA57147.2024.10611477}
}

@InProceedings{pmlr-v270-kim25c,
  title = {{OpenVLA}: An open-source vision-language-action model},
  author = {Kim, Moo Jin and Pertsch, Karl and Karamcheti, Siddharth and Xiao, Ted and Balakrishna, Ashwin and Nair, Suraj and Rafailov, Rafael and Foster, Ethan P and Sanketi, Pannag R and Vuong, Quan and Kollar, Thomas and Burchfiel, Benjamin and Tedrake, Russ and Sadigh, Dorsa and Levine, Sergey and Liang, Percy and Finn, Chelsea},
  booktitle = {Proceedings of The 8th Conference on Robot Learning},
  pages = {2679--2713},
  year = {2025},
  volume = {270},
  series = {Proceedings of Machine Learning Research},
  publisher = {PMLR}
}

@InProceedings{Zhang_2025_ICCV,
  author = {Zhang, Shiduo and Xu, Zhe and Liu, Peiju and Yu, Xiaopeng and Li, Yuan and Gao, Qinghui and Fei, Zhaoye and Yin, Zhangyue and Wu, Zuxuan and Jiang, Yu-Gang and Qiu, Xipeng},
  title = {{VLABench}: A large-scale benchmark for language-conditioned robotics manipulation with long-horizon reasoning tasks},
  booktitle = {Proceedings of the IEEE/CVF International Conference on Computer Vision (ICCV)},
  month = {October},
  year = {2025},
  pages = {11142--11152}
}

@InProceedings{pmlr-v305-zhao25a,
  title = {{ManipBench}: Benchmarking vision-language models for low-level robot manipulation},
  author = {Zhao, Enyu and Raval, Vedant and Zhang, Hejia and Mao, Jiageng and Shangguan, Zeyu and Nikolaidis, Stefanos and Wang, Yue and Seita, Daniel},
  booktitle = {Proceedings of The 9th Conference on Robot Learning},
  pages = {3413--3462},
  year = {2025},
  volume = {305},
  series = {Proceedings of Machine Learning Research},
  publisher = {PMLR}
}

@inproceedings{quartey2025verifiably,
  title = {Verifiably following complex robot instructions with foundation models},
  author = {Quartey, Benedict and Rosen, Eric and Tellex, Stefanie and Konidaris, George},
  booktitle = {2025 IEEE International Conference on Robotics and Automation (ICRA)},
  pages = {1--8},
  year = {2025},
  organization = {IEEE},
  doi = {10.1109/ICRA55743.2025.11127418}
}

@inproceedings{chiyah-garcia-etal-2024-repairs,
  title = {Repairs in a {Block World}: A new benchmark for handling user corrections with multi-modal language models},
  author = {Chiyah-Garcia, Javier and Suglia, Alessandro and Eshghi, Arash},
  booktitle = {Proceedings of the 2024 Conference on Empirical Methods in Natural Language Processing},
  month = nov,
  year = {2024},
  address = {Miami, Florida, USA},
  publisher = {Association for Computational Linguistics},
  doi = {10.18653/v1/2024.emnlp-main.643},
  pages = {11523--11542}
}

@inproceedings{stoyanchev-etal-2026-context,
  title = {Context-aware language understanding in human-robot dialogue with {LLMs}},
  author = {Stoyanchev, Svetlana and Farag, Youmna and Keizer, Simon and Li, Mohan and Doddipatla, Rama Sanand},
  booktitle = {Proceedings of the 16th International Workshop on Spoken Dialogue System Technology},
  month = feb,
  year = {2026},
  address = {Trento, Italy},
  publisher = {Association for Computational Linguistics},
  pages = {262--274}
}

@misc{li2025switchvlaexecutionawaretaskswitching,
  title = {{SwitchVLA}: Execution-aware task switching for vision-language-action models},
  author = {Li, Meng and Zhao, Zhen and Che, Zhengping and Liao, Fei and Wu, Kun and Xu, Zhiyuan and Ren, Pei and Jin, Zhao and Liu, Ning and Tang, Jian},
  year = {2025},
  eprint = {2506.03574},
  archivePrefix = {arXiv},
  primaryClass = {cs.RO},
  note = {arXiv:2506.03574}
}

@inproceedings{oba-sugawara-2026-cxmp,
  title = {{CxMP}: A linguistic minimal-pair benchmark for evaluating constructional understanding in language models},
  author = {Oba, Miyu and Sugawara, Saku},
  booktitle = {Proceedings of the 64th Annual Meeting of the Association for Computational Linguistics (Volume 1: Long Papers)},
  month = jul,
  year = {2026},
  address = {San Diego, California, United States},
  publisher = {Association for Computational Linguistics},
  doi = {10.18653/v1/2026.acl-long.2132},
  pages = {45949--45963},
  isbn = {979-8-89176-390-6}
}

@misc{kim2026liberoparadiagnosticbenchmarkmetrics,
  title = {{LIBERO-Para}: A diagnostic benchmark and metrics for paraphrase robustness in {VLA} models},
  author = {Kim, Chanyoung and Kim, Minwoo and Kang, Minseok and Kim, Hyunwoo and Jung, Dahuin},
  year = {2026},
  eprint = {2603.28301},
  archivePrefix = {arXiv},
  primaryClass = {cs.LG},
  note = {arXiv:2603.28301}
}

@article{mesa2026,
  title = {{MESA}: An evaluation framework for compositional, semantic, and spatial generalization in robotics},
  author = {Wilcox, Albert and Chang, Frank and Chakraborty, Aishani and Nguyen, Nhi and Collins, Jeremy A. and Saxena, Vaibhav and Joffe, Benjamin and Karamcheti, Siddharth and Garg, Animesh},
  journal = {arXiv preprint},
  year = {2026}
}

@misc{qi2026scalestrategicallylearningcompositional,
  title = {Scale up strategically: Learning compositional generalization via bias-aware evaluation and data collection for robotic manipulation},
  author = {Qi, Yu and Ye, Zhang and Xu, Xinyi and Lu, Yuxuan and Sandhu, Amitoj and Hu, Boce and Huang, Haojie and Tremblay, Jonathan and Wong, Lawson L. S.},
  year = {2026},
  eprint = {2607.21582},
  archivePrefix = {arXiv},
  primaryClass = {cs.RO},
  note = {arXiv:2607.21582}
}

@misc{wu2026atombench,
  title = {{ATOM-Bench}: A real-world benchmark for atomic skills and compositional generalization in manipulation policies},
  author = {Wu, Zenan and Wei, Bingqing and Liu, Lu and He, Zheqi and Wang, Xi and Liu, Jiakang and Li, Zehui and Yao, Guocai and Zheng, Jing-Shu and Yang, Xi and Wang, Yongtao},
  year = {2026},
  eprint = {2606.16826},
  archivePrefix = {arXiv},
  primaryClass = {cs.RO},
  note = {arXiv:2606.16826}
}

@misc{wang2026diagnosingcompositional,
  title = {Diagnosing compositional generalization in sequential robot tasks},
  author = {Wang, Yixiao and Wu, Cheng-En and Sun, Lingfeng and Wang, Pengcheng and Ji, Xiang and Liang, Boyuan and Zhan, Guojian and Tomizuka, Masayoshi},
  year = {2026},
  eprint = {2607.29687},
  archivePrefix = {arXiv},
  primaryClass = {cs.RO},
  note = {arXiv:2607.29687}
}

@misc{glm5team2026glm5vibecodingagentic,
  title = {{GLM-5}: From vibe coding to agentic engineering},
  author = {{GLM-5-Team} and Aohan Zeng and Xin Lv and Zhenyu Hou and Zhengxiao Du and Qinkai Zheng and Bin Chen and Da Yin and Chendi Ge and Chenghua Huang and Chengxing Xie and Chenzheng Zhu and Congfeng Yin and Cunxiang Wang and Gengzheng Pan and Hao Zeng and Haoke Zhang and Haoran Wang and Huilong Chen and Jiajie Zhang and Jian Jiao and Jiaqi Guo and Jingsen Wang and Jingzhao Du and Jinzhu Wu and Kedong Wang and Lei Li and Lin Fan and Lucen Zhong and Mingdao Liu and Mingming Zhao and Pengfan Du and Qian Dong and Rui Lu and Shuang-Li and Shulin Cao and Song Liu and Ting Jiang and Xiaodong Chen and Xiaohan Zhang and Xuancheng Huang and Xuezhen Dong and Yabo Xu and Yao Wei and Yifan An and Yilin Niu and Yitong Zhu and Yuanhao Wen and Yukuo Cen and Yushi Bai and Zhongpei Qiao and Zihan Wang and Zikang Wang and Zilin Zhu and Ziqiang Liu and Zixuan Li and Bojie Wang and Bosi Wen and Can Huang and Changpeng Cai and Chao Yu and Chen Li and Chengwei Hu and Chenhui Zhang and Dan Zhang and Daoyan Lin and Dayong Yang and Di Wang and Ding Ai and Erle Zhu and Fangzhou Yi and Feiyu Chen and Guohong Wen and Hailong Sun and Haisha Zhao and Haiyi Hu and Hanchen Zhang and Hanrui Liu and Hanyu Zhang and Hao Peng and Hao Tai and Haobo Zhang and He Liu and Hongwei Wang and Hongxi Yan and Hongyu Ge and Huan Liu and Huanpeng Chu and Jia'ni Zhao and Jiachen Wang and Jiajing Zhao and Jiamin Ren and Jiapeng Wang and Jiaxin Zhang and Jiayi Gui and Jiayue Zhao and Jijie Li and Jing An and Jing Li and Jingwei Yuan and Jinhua Du and Jinxin Liu and Junkai Zhi and Junwen Duan and Kaiyue Zhou and Kangjian Wei and Ke Wang and Keyun Luo and Laiqiang Zhang and Leigang Sha and Liang Xu and Lindong Wu and Lintao Ding and Lu Chen and Minghao Li and Nianyi Lin and Pan Ta and Qiang Zou and Rongjun Song and Ruiqi Yang and Shangqing Tu and Shangtong Yang and Shaoxiang Wu and Shengyan Zhang and Shijie Li and Shuang Li and Shuyi Fan and Wei Qin and Wei Tian and Weining Zhang and Wenbo Yu and Wenjie Liang and Xiang Kuang and Xiangmeng Cheng and Xiangyang Li and Xiaoquan Yan and Xiaowei Hu and Xiaoying Ling and Xing Fan and Xingye Xia and Xinyuan Zhang and Xinze Zhang and Xirui Pan and Xu Zou and Xunkai Zhang and Yadi Liu and Yandong Wu and Yanfu Li and Yidong Wang and Yifan Zhu and Yijun Tan and Yilin Zhou and Yiming Pan and Ying Zhang and Yinpei Su and Yipeng Geng and Yong Yan and Yonglin Tan and Yuean Bi and Yuhan Shen and Yuhao Yang and Yujiang Li and Yunan Liu and Yunqing Wang and Yuntao Li and Yurong Wu and Yutao Zhang and Yuxi Duan and Yuxuan Zhang and Zezhen Liu and Zhengtao Jiang and Zhenhe Yan and Zheyu Zhang and Zhixiang Wei and Zhuo Chen and Zhuoer Feng and Zijun Yao and Ziwei Chai and Ziyuan Wang and Zuzhou Zhang and Bin Xu and Minlie Huang and Hongning Wang and Juanzi Li and Yuxiao Dong and Jie Tang},
  year = {2026},
  eprint = {2602.15763},
  archivePrefix = {arXiv},
  primaryClass = {cs.LG},
  note = {arXiv:2602.15763}
}

@misc{qwen2025qwen25technicalreport,
  title = {{Qwen2.5} technical report},
  author = {{Qwen} and An Yang and Baosong Yang and Beichen Zhang and Binyuan Hui and Bo Zheng and Bowen Yu and Chengyuan Li and Dayiheng Liu and Fei Huang and Haoran Wei and Huan Lin and Jian Yang and Jianhong Tu and Jianwei Zhang and Jianxin Yang and Jiaxi Yang and Jingren Zhou and Junyang Lin and Kai Dang and Keming Lu and Keqin Bao and Kexin Yang and Le Yu and Mei Li and Mingfeng Xue and Pei Zhang and Qin Zhu and Rui Men and Runji Lin and Tianhao Li and Tianyi Tang and Tingyu Xia and Xingzhang Ren and Xuancheng Ren and Yang Fan and Yang Su and Yichang Zhang and Yu Wan and Yuqiong Liu and Zeyu Cui and Zhenru Zhang and Zihan Qiu},
  year = {2025},
  eprint = {2412.15115},
  archivePrefix = {arXiv},
  primaryClass = {cs.CL},
  note = {arXiv:2412.15115}
}

@misc{mistralai2024mistral7binstructv03,
  author = {{Mistral AI}},
  title = {{Mistral-7B-Instruct-v0.3}},
  year = {2024},
  note = {[Online]. Available: \url{https://huggingface.co/mistralai/Mistral-7B-Instruct-v0.3}}
}

@misc{teamolmo2025olmo2furious,
  title = {2 {OLMo} 2 furious},
  author = {{Team OLMo} and Pete Walsh and Luca Soldaini and Dirk Groeneveld and Kyle Lo and Shane Arora and Akshita Bhagia and Yuling Gu and Shengyi Huang and Matt Jordan and Nathan Lambert and Dustin Schwenk and Oyvind Tafjord and Taira Anderson and David Atkinson and Faeze Brahman and Christopher Clark and Pradeep Dasigi and Nouha Dziri and Allyson Ettinger and Michal Guerquin and David Heineman and Hamish Ivison and Pang Wei Koh and Jiacheng Liu and Saumya Malik and William Merrill and Lester James V. Miranda and Jacob Morrison and Tyler Murray and Crystal Nam and Jake Poznanski and Valentina Pyatkin and Aman Rangapur and Michael Schmitz and Sam Skjonsberg and David Wadden and Christopher Wilhelm and Michael Wilson and Luke Zettlemoyer and Ali Farhadi and Noah A. Smith and Hannaneh Hajishirzi},
  year = {2025},
  eprint = {2501.00656},
  archivePrefix = {arXiv},
  primaryClass = {cs.CL},
  note = {arXiv:2501.00656}
}

\end{document}